\documentclass[letterpaper, 10 pt, journal, twoside]{IEEEtran}  % Comment this line out if you need a4paper

\usepackage{algorithm}
\usepackage[noend]{algpseudocode}
\usepackage{varwidth} % For maintaining states at the same indentation
\usepackage{caption} %- Caption hacks
\usepackage{multirow}
\usepackage{hhline}
\usepackage{color,soul} % Highlighting

\newcommand{\larw}{$\leftarrow$\ }

\makeatletter % Algo package hack
\def\BState{\State\hskip-\ALG@thistlm}
\makeatother

\newcommand\CONDITION[2]%
{\begin{tabular}[t]{@{}l@{}l@{}}
		#1&#2
	\end{tabular}%
}
\algdef{SE}[WHILE]{While}{EndWhile}[1]%
{\algorithmicwhile\ \CONDITION{#1}{\ \algorithmicdo}}%
{\algorithmicend\ \algorithmicwhile}
\algdef{SE}[FOR]{For}{EndFor}[1]%
{\algorithmicfor\ \CONDITION{#1}{\ \algorithmicdo}}%
{\algorithmicend\ \algorithmicfor}
\algdef{S}[FOR]{ForAll}[1]%
{\algorithmicforall\ \CONDITION{#1}{\ \algorithmicdo}}
\algdef{SE}[REPEAT]{Repeat}{Until}{\algorithmicrepeat}[1]%
{\algorithmicuntil\ \CONDITION{#1}{}}
\algdef{SE}[IF]{If}{EndIf}[1]%
{\algorithmicif\ \CONDITION{#1}{\ \algorithmicthen}}%
{\algorithmicend\ \algorithmicif}%
\algdef{C}[IF]{IF}{ElsIf}[1]%
{\algorithmicelse\ \algorithmicif\ \CONDITION{#1}{\ \algorithmicthen}}

\newcommand{\subparagraph}{}
\usepackage{graphicx} % for pdf, bitmapped graphics files
\usepackage{float} % -- Forcing figure position
\usepackage{amsmath} % assumes amsmath package installed
\usepackage{amssymb}  % assumes amsmath package installed

\usepackage{xfrac}
\usepackage{cite} % Group multiple citations into brackets

\title{Assembly Planning by Subassembly Decomposition \\Using Blocking Reduction}

\author{James Watson$^{1}$ and Tucker Hermans$^{2}$% <-this % stops a space
\thanks{Manuscript received: March 5, 2019; Revised May 17, 2019; Accepted June 27, 2019. This paper was recommended for publication by Dan Popa upon evaluation of the Associate Editor and Reviewers' comments.}%Use only for final RAL version
\thanks{$^{1}$Dept. of Computer Science,
	University of Colorado Boulder (completed at U.\! of\! Utah),
	Boulder, Colorado 80303,
    {\tt\footnotesize james.watson-2@colorado.edu}}
\thanks{$^{2}$School of Computing,
	University of Utah,
	Salt Lake City, Utah 84112,
    {\tt\footnotesize thermans@cs.utah.edu}}%
\thanks{Digital Object Identifier (DOI): see top of this page.}
}

\begin{document}

\maketitle

%%%%%%%%%%%%%%%%%%%%%%%%%%%%%%%%%%%%%%%%%%%%%%%%%%%%%%%%%%%%%%%%%%%%%%%%%%%%%%%%
\begin{abstract}

The sequence in which a complex product is assembled directly impacts the ease and efficiency of the assembly process, whether executed by a human or a robot. A sequence that gives the assembler the greatest freedom of movement is therefore desirable.  Our main contribution is an expression of obstruction relationships between parts as a disassembly interference graph (DIG).  We validate this heuristic by developing a disassembly sequence planner that partitions assemblies in a way that prioritizes access to parts, resulting in plans that are comparable in efficiency to two state-of-the-art assembly methods in terms of total plan length.  Using DIG, our method generates successive subassembly decompositions, yielding a tree structure that makes parallization opportunities apparent.  Our planner generates viable disassembly plans by minimizing our part blockage measure, and thereby demonstrates that this measure is a valuable addition to the Assembly Sequence Planning toolkit.

\end{abstract}
\def \slideScale {0.40}
%%%%%%%%%%%%%%%%%%%%%%%%%%%%%%%%%%%%%%%%%%%%%%%%%%%%%%%%%%%%%%%%%%%%%%%%%%%%%%%% 
\section{Introduction}

\IEEEPARstart{A}{ssembly} planning concerns the ordering and planning of actions that bring separate parts together to form a complete, complex product assembly \cite{simplifiedGenAllSeq_GEARTRAIN}.  The study of assembly planning is important to both the human performance of assembly on the factory floor \cite{PostProcessingRafiCampbell}, and to the development of fully autonomous industrial robotics systems \cite{plansForRobotExec}.  Assembly Sequence Planning (ASP) is a subproblem of assembly planning that focuses on establishing the order in which to assemble parts, given some suitable planner for the individual motions \cite{simplifiedGenAllSeq_GEARTRAIN}.  

The identification of subassemblies within a product is central to assembly planning.  We hypothesize that partitioning a product into subassemblies in a way that respects both assembly constraints and part access will result in assembly plans that are time-efficient in the number of actions required to execute the generated plans.  Existing assembly planning methods deal with part access as a separate validation step \cite{IKEA,kmeansRRT45,volumeAreaFitness2}.  Our method addresses part access as a direct, quantitative measure.

Our main contribution is a graph structure which estimates the interference that each part presents to the disassembly of all other parts, called the disassembly interference graph (DIG).  We exploit this graph to effectively partition complex assemblies into logical subassemblies that respect correct assembly precedence and yield disassembly sequence plans with short makespans.  Our approach has the advantage of treating part disassembly interference as a cost to reduce during subassembly identification.  This prevents our planner from rejecting advantageous subassemblies during the formation phase.  In the state of the art, and historically, part interference is always treated as a hard constraint \cite{liaisonModern,validationBlock_kheder2017,clumpNDBG}.  

To show our method's efficacy, we compare it to two other subassembly partition methods \cite{kmeansRRT45,volumeAreaFitness2}.  We evaluate performance in terms of the number of actions required to execute the resulting sequences.  The other methods focus on geometric properties of parts and their contacts and degree of liaison graph nodes \cite{volumeAreaFitness,volumeAreaFitness2}, and geometric proximity \cite{kmeansRRT45}.  The comparative methods use motion planning to validate subassembly choices, but do not consider part interaction as part of the subassembly choice, as ours does.

\section{Related Work}

\subsection{Subassembly Identification}

The enormous combinatoric size of possible assembly sequences for any product motivates subassembly identification. There are $ ( 2n - 2 )!/( n - 1 )! $ possible sequences of actions for an assembly with $n$ parts, if it is assumed that assembly actions are monotonic \cite{AsmSizeComplexity}.  This includes both feasible and infeasible sequences. Partitioning a complete product assembly into sub-assemblies is an effective strategy for containing the combinatoric explosion.  This partitioning task is itself an NP-Complete problem \cite{subComplexity}.  

A straightforward approach to subassembly identification is to group parts by proximity, as Morato et al. did in \cite{kmeansRRT45}.  Belhadj et al. \cite{volumeAreaFitness} identify parts with many connections, large surface area, and large volume as candidates to support nearby parts in a sub-assembly. They build on this work in \cite{volumeAreaFitness2} by requiring that parts assigned to the subassembly have a stronger connection to the subassembly than to the remainder of the assembly.  Lee et al. \cite{forceFlowDisasm} focus on the forces required to separate a subassembly and define subassemblies as the grouping that relieves the most disassembly forces at once, found by max-flow min-cut \cite{maxFlowMinCut}.  Vigan{\`o} and G{\'o}mez speed up their search by delaying evaluation of feasibility until after sub-assemblies are identified. \cite{CollsnCheckSubsOnly}
\subsection{ASP with Motion Planning}
Current methods seek to incorporate motion planning into ASP.  That is, the cost and quality of motion plans influences the generation of sequences.  The goal of these methods is to take into account the difficulty of the task-space execution of assembly sequences.  Wan and Harada \cite{SingleArmAsmRegraspWanHarada} incorporate grasp planning and motion planning into each assembly operation, executed with a serial manipulator. Morato et al. \cite{kmeansRRT45} generate full motion plans to validate candidate subassemblies for their ability to be installed into the product assembly as a unit. The method we present fits into the hybrid category, as simple motion plans are generated as a validation step.

This work does not address generating part fixtures for assembly, but this is an area of ongoing research. \cite{SandiaModularFixture,FxtrSocialAutoBody,SupportOnPinsWanHarada}
\section{Method}
The basis for our method is the intuition that the ability to execute an assembly depends on access to individual parts.  Part access can be modeled using the assembly geometry alone.  Following this intuition, we have developed a method that identifies subassemblies in a way that prioritizes part access.  Part access can also be used to establish precedence among identified subassemblies.

This section is structured as follows. We define the concept of removal spaces as a necessary consideration for assembly planning.  Blocking fraction is introduced as a new heuristic to measure the impingement upon removal spaces.  Based on this, we develop the disassembly influence graph (DIG) as a structure that expresses how each part blocks the other from being removed. Finally, we propose an assembly sequence planner that exploits part access as the primary consideration for sequence generation.
\def \figScale {0.185} 
\begin{figure}[t]
	\centering
	\includegraphics[scale=\figScale]{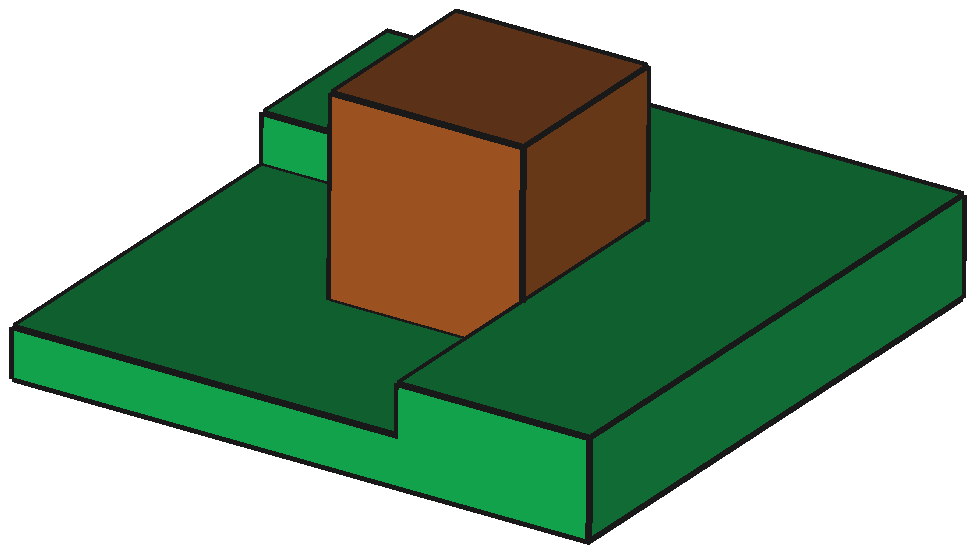} 
	\caption{A cube rests in a notch cut from a cuboid plate. The cube shown sits in a recess that allows for positive translation along the three principal axes, or any direction in between.} \label{simpleCube}
	\vspace{-3.5mm}
\end{figure}
\begin{figure}[b]
		\centering
		\includegraphics[scale=\figScale]{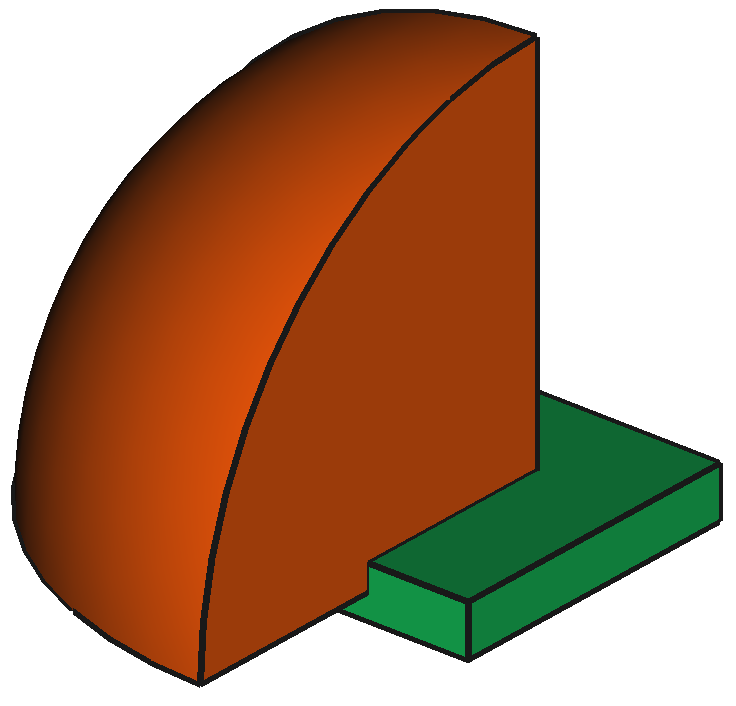} 
		\caption{Removal space of the cube shown in Figure \ref{simpleCube}, represented by a spherical pyramid. Its flat surfaces are defined by the constraints on local freedom.}
		\label{removalSpace}
\end{figure}
\subsection{Removal Space} \label{SECT_removalSpace}

Defining the removal space is the first step in quantifying part access.  In order to model part access, we define a 3-dimensional volume for each part called a removal space.  Local movement constraints between parts imply a removal space representing the region of translation removal paths.  Once this region is defined, it can be evaluated for obstructions that might impair disassembly actions, without generating and evaluating individual paths.  Using the model, part access, or conversely part blockage, can be measured as a scalar quantity.

The removal space models a continuous space of possible translations of a part away from it's specified relative pose within the complete assembly.  Consider a cube resting on a plate with a notch cut out of it. (Figure \ref{simpleCube})  It may be translated away from the plate vertically,  horizontally along either of the notch walls, or any direction that has a positive dot product with all of the notch face normals.  These translation constraints may be visualized as the spherical pyramid in Figure \ref{removalSpace}.  The removal space for a part is a region that begins at the part's assembled location and radiates out along possible translation paths as it exits the assembly.  We can imagine the curved outward-facing surface of this region as a boundary (shell) through which the cube must translate to be removed.
\begin{figure}[b]
	\vspace{-3.5mm}
	\centering
	\includegraphics[scale=0.15]{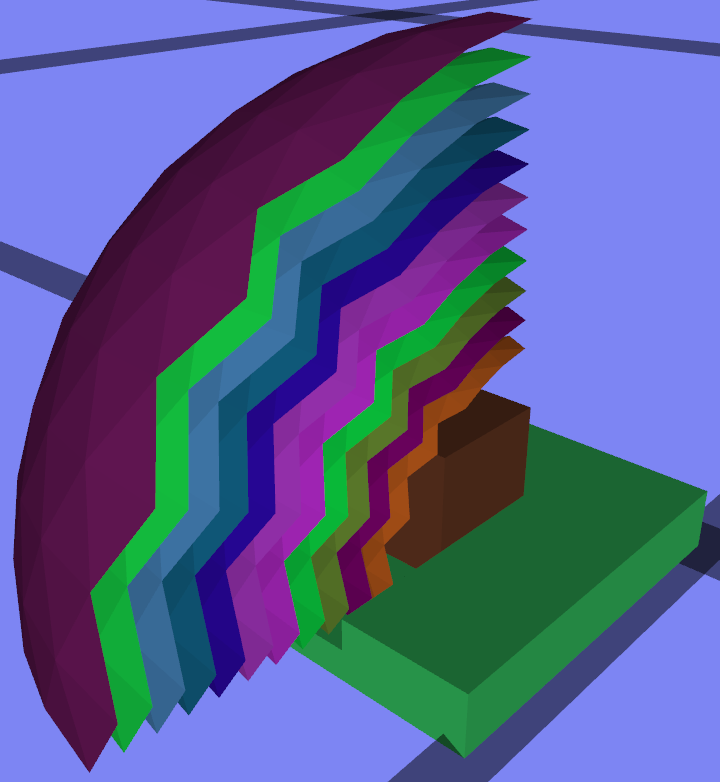}
	\caption{Intersection surfaces, Spherical Pyramid. }
	\label{sphereShell}
\end{figure}
\subsection{Blocking Fraction} \label{SECT_disasmInterfere}
We provide a heuristic for estimating the degree to which one Part $j$ obstructs another Part $i$ from being disassembled, called the blocking fraction ($w_{ij}$).  We define the blocking fraction $w_{ij}$ as an expression of the degree to which the removal space of a Part $i$ is obstructed by another Part $j$.  Blocking fraction may be thought of as the complement of part access.

The first step of calculating blocking fraction is the construction of surfaces (shells), all concentric to the outer boundary of the removal space. (Figure \ref{sphereShell})  At any distance $d$ from the part centroid, regions of the corresponding shell that are in intersection with surrounding assembly geometry represent positions unavailable for the part to occupy on its way out.  The greater the intersection area, the greater the obstruction.  The maximum fraction of the surface area of any shell that is in collision with another part's geometry is the blocking fraction.

The spherical pyramid representation of free removal space is not new in ASP.  Romney analyzes the geometry of assemblies to generate an extended translational freedom cone in \cite{freedomCone}.  Thomas et al. \cite{projectionCone2p5D} use a cone in configuration space to determine the feasibility of part removal operations.   Our method is different from the usual in that we compute the obstruction of a cone of candidate removal directions, and treat this blockage as a cost to be reduced.

Calculation of the blocking fraction for part $P_i$ vs $P_j , i \neq j$ proceeds in three steps: determination of freedom type from local part constraints, generation of concentric surfaces, and calculating intersection of parts and surfaces.  The constraints imposed by adjacent parts $P_j$ (reference assembly) determine the degrees of freedom of $P_i$ (moving assembly).  Our system can identify several types of freedom scenario ranging from zero to three degrees of translational freedom.
\begin{algorithm}
	\caption{Blocking Fraction, with Moved Part $P_{i}$ and Reference Part $P_{j}$} \label{calcBlockingFrac}
	\begin{algorithmic}[1]
		\Procedure{Blocking Fraction}{$P_{i}$, $P_{j}$, NDBG}
		
		\begin{flushleft}
			\textbf{INPUT:} Part $P_{i}$, Part $P_{j}$, NDBG \\
			\textbf{OUTPUT:} Blocking Fraction $w_{ij} \in \left[0,1\right]$
		\end{flushleft}
		
		\State $w_{ij} = 0$
		\State {\tt shells = construct\_shells(} $P_i$ , NDBG {\tt )}
		\For{ {\tt shell $s$ in shells} }
		\State $A_{block}$ \larw {\tt mesh\_intersection(} $s$ , $P_j$ {\tt )}
		\State $A_{s}$ \larw {\tt area\_of(} $s$ {\tt )}
		\State $w_{ij}$ \larw $\max( w_{ij} , \sfrac{A_{block}}{A_s} )$
		\EndFor
		\State \Return $w_{ij}$
		\EndProcedure
	\end{algorithmic}
\end{algorithm}
Construction of intersection surfaces begins at the volume centroid of the part, and the intervals between surfaces must be spaced sufficiently close together to correctly capture when one part passes through the interior of another. 

\subsection{Disassembly Influence Graph} \label{SECT_DIGtotalBlock}

We define the disassembly influence graph (DIG) as a means of describing the degree to which each part blocks every other part from being removed from the product assembly.  The vertices $P_D$ represent all of the parts in the product assembly in their relative poses in the completed product, and directed edges $e(P_j,P_i) \in E_D$ represent the degree to which $P_i$ is blocked from removal by $P_j$.  The weight $w_{ij} \in W_D$ represents the blocking fraction $P_i$ that intersects $P_j$.  The DIG has $n^2$ weights $w_{ij}$, and can be represented by an $n \times n$ matrix.  In general, this matrix is not symmetric.  The interference measure can also be calculated for collections of subassemblies.

The blocking fraction describes the blocked relationship between $P_i$ and $P_j$.  The total blockage $\tau_i$ for part $i$ describes the degree to which a single part or subassembly is blocked by all other parts, and is defined by:  
\begin{align}
\tau_i = \sum_{j=1}^{N}\left[ w_{i,j} \right]
\end{align}
\noindent where $N$ is the total number of parts in the assembly considered. The quantity $\tau_i$ can be greater than one, which can happen either when the part $i$ has many partial blockages along its possible exit paths, or when it is completely covered by more than one part.

Computation of the DIG requires a number of collision detection calculations; $O(mn^2)$, where $n$ is the number of parts in the subassembly, and $m$ is the number of concentric shells used in each part-part interference determination. It is only necessary to calculate the DIG once for each level of disassembly, and only for the parts present in that subassembly.  Part interference does not need to be recalculated during iterative subassembly identification for different choices of nucleus parts. Our method also benefits from the efficiency of the RAPID collision detection system \cite{RAPID}, which drastically reduces triangle-triangle intersection tests necessary for mesh collision checking through the use of hierarchical, object-aligned bounding boxes that are assigned before planning begins.

\subsection{Subassembly Identification by Blocking Reduction} \label{SIR}

Our method decomposes a complete assembly into subassemblies whenever possible, and only falls back on single-part removal operations when no such groupings can be made.  There are two main benefits to subdividing the disassembly problem.  The first is that it allows the problem to be solved recursively, with each subassembly being treated as a smaller disassembly problem.  The second is that independent subassemblies may be executed in parallel.
\begin{algorithm}
	\caption{Subassembly Identification by Blocking Reduction} \label{blockingReduceAlgo}
	\begin{algorithmic}[1]
		
		\Procedure{SubID by Blocking Reduction}{Assembly State S}
		
		\begin{flushleft}
			\textbf{INPUT:} Assembly part geometry, relative poses, liaison graph, NDBG \\
			\textbf{OUTPUT:} List of subassemblies
		\end{flushleft}
		
		\State Identify base
		\State Identify nucleus parts
		\State Calculate DIG using nucleus-base contact
		\For{Each nucleus part} 
		\State Accrue neighbors that do not increase blockage of subassembly
		\State Consider neighbors of added part
		\EndFor
		\State Accept subassemblies below blocking threshold
		
		\For{For each accepted subassembly}
		\State Validate subassembly removal
		\EndFor
		
		\State \Return Validated subassemblies
		
		\EndProcedure
	\end{algorithmic}
\end{algorithm}

Generating a disassembly plan that prioritizes part access proceeds along the following steps; choosing candidate subassembly bases, determining removal spaces for parts of the assembly, composing subassemblies with removal spaces free from interference, and executing removal actions with the freest part access first.

The primary aim of our method is to pull parts away from the total assembly in groups that have the freest removal space.  This is done by first choosing candidate subassembly centers (nuclei) based on a global heuristic (as in \cite{volumeAreaFitness}), and then growing each candidate subassembly by conditionally adding adjacent parts that minimize a measure of access we have developed, called blocking fraction.

\begin{figure}[h!]
	\centering
	\includegraphics[scale=0.25]{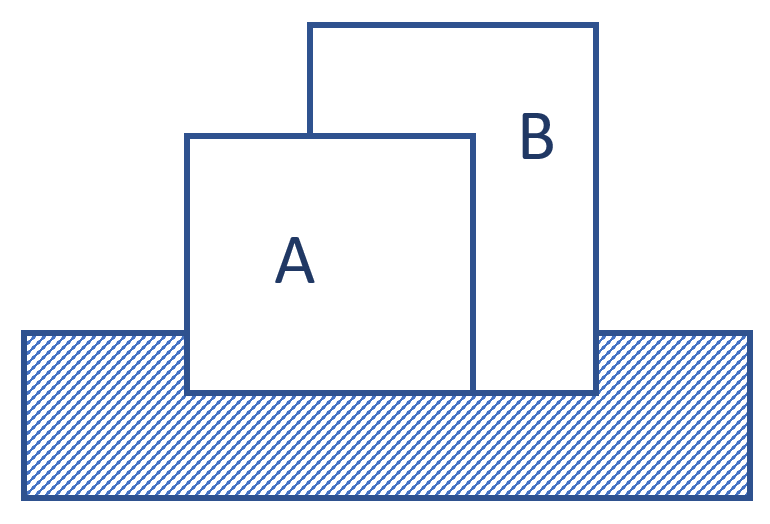}
	\caption{Part $A$ cannot be removed alone, but a subassembly consisting of $A$ and $B$ can.}
	\label{A_accrues_B}
	\vspace{-1.5mm}
\end{figure}

A single part may be partially blocked by several neighbors, and cannot be removed.  However, if this part can be removed with its blocking neighbors as a unit, then it is a logical candidate for a subassembly.  (Figure \ref{A_accrues_B})  Also, adjacent parts that can be removed as a unit become part of a subassembly.  Individual parts accrue neighbors whose addition does not increase the total blockage $\tau_{sub}$ of a candidate subassembly.  After candidate subassemblies are generated, we calculate the total blockage for each as a unit.  Subassemblies that have a sufficiently free removal space are validated by simulation.

The first step in the subassembly identification process is to determine a base part.  The part with the most volume serves as the base part for the complete assembly, and is removed from consideration for subassemblies, as in \cite{kmeansRRT45}.  The remaining parts are graded for their fitness as the nuclei of subassemblies using the fitness score developed by Belhadj \cite{volumeAreaFitness}.  The largest gap in score between any two part scores in a sorted, descending list is chosen as the cutoff between nuclei parts and all others.  Then, the DIG is calculated for all non-base parts (including nuclei), using only base-part constraints to determine the local freedom of each part.  The DIG is calculated using this freest state so that changes in blockage can be evaluated with the addition of each part to an assembly.  

Formation of the first subassembly begins by considering the nucleus part with the greatest base part fitness score. All of the immediate neighbors of the nucleus part are added to a priority queue, sorted by the degree to which each blocks the nucleus part (blocking fraction).  The total blockage $\tau_{sub}$ for an assembly composed of the nucleus part and the prospective addition is calculated by summing the rows of the DIG that represent the participating parts, while excluding the columns that correspond to those parts. If the addition of the part frees a locked subassembly, the addition is automatically accepted. (Figure \ref{A_accrues_B})  Likewise, an addition that causes a subassembly to be locked is automatically rejected.  If there was no change to or from a locked state, then a prospective part is accepted into a subassembly if the total blockage of the subassembly is less than or equal to the blockage without the new part.  If a part is accepted, then its neighbors are enqueued.  
The next most-blocking part is then popped from the queue, and the process repeats on parts that have neither been evaluated nor assigned to another subassembly.  Once the queue is emptied, the subassembly accumulation process is repeated on the nucleus part with the next-highest score.  The identification phase ends when all nuclei and their neighbors have been evaluated as described.  Any non-base parts that remain unassigned at the end of the identification phase are grouped into a single remainder subassembly that includes the base.

The total blockage of each of the identified subassemblies is evaluated with respect to the entire assembly.  Only subassemblies with a total blockage below a given threshold $\tau_{sub} \leq f_{accept}$ are considered for removal.  This prevents us from simulating removals that are likely to fail.  We empirically found $f_{accept}=0.85$ to work well in our experiments.  Simulated removal is attempted for the subassemblies that meet the criterion, in directions sampled from the subassembly local freedom.  

If the subassembly is able to be removed from the complete assembly, then it is added to the precedence layer.  Unsuccessful subassemblies are added back onto a remainder subassembly to be decomposed at the next recursion.

\subsection{Sequence Generation from Disassembly Tree} \label{sequence}

The result of subassembly identification at each recursion depth is a precedence layer \cite{kmeansRRT45} that represents all of the disassembly operations that can be done without respect to order.  Successive decomposition of the assembly into precedence layers results in a tree structure that represents the parallelism of the (dis)assembly process in a straightforward way.  This tree structure can be used to represent the output of all three subassembly identification methods described here. 

By design, disassembly actions associated with sibling nodes in the precedence layer do not interfere with each other and have no precedence over another.  Sibling actions can be performed in any order without affecting makespan.  %It is possible to perform disassembly actions within the same precedence layer simultaneously \cite{HANDEY}, but this is a separate planning problem outside the scope of this work.

An assembly sequence can be readily derived from the disassembly tree by starting with any leaf node and performing the join operation represented by the edge from the node to the parent, then proceeding with that leaf's siblings. Leaves associated with completed actions are removed and a new leaf can be chosen for the next action.  A parent subassembly cannot be joined to the grandparent subassembly until all the child join operations have been completed. The disassembly tree also encodes the parallel nature of a plan; any two subtrees with only leaf successors can be assigned to two different workers.

\def \caseScale {0.15}
\begin{figure}[b]
	\centering
	\includegraphics[scale=\caseScale]{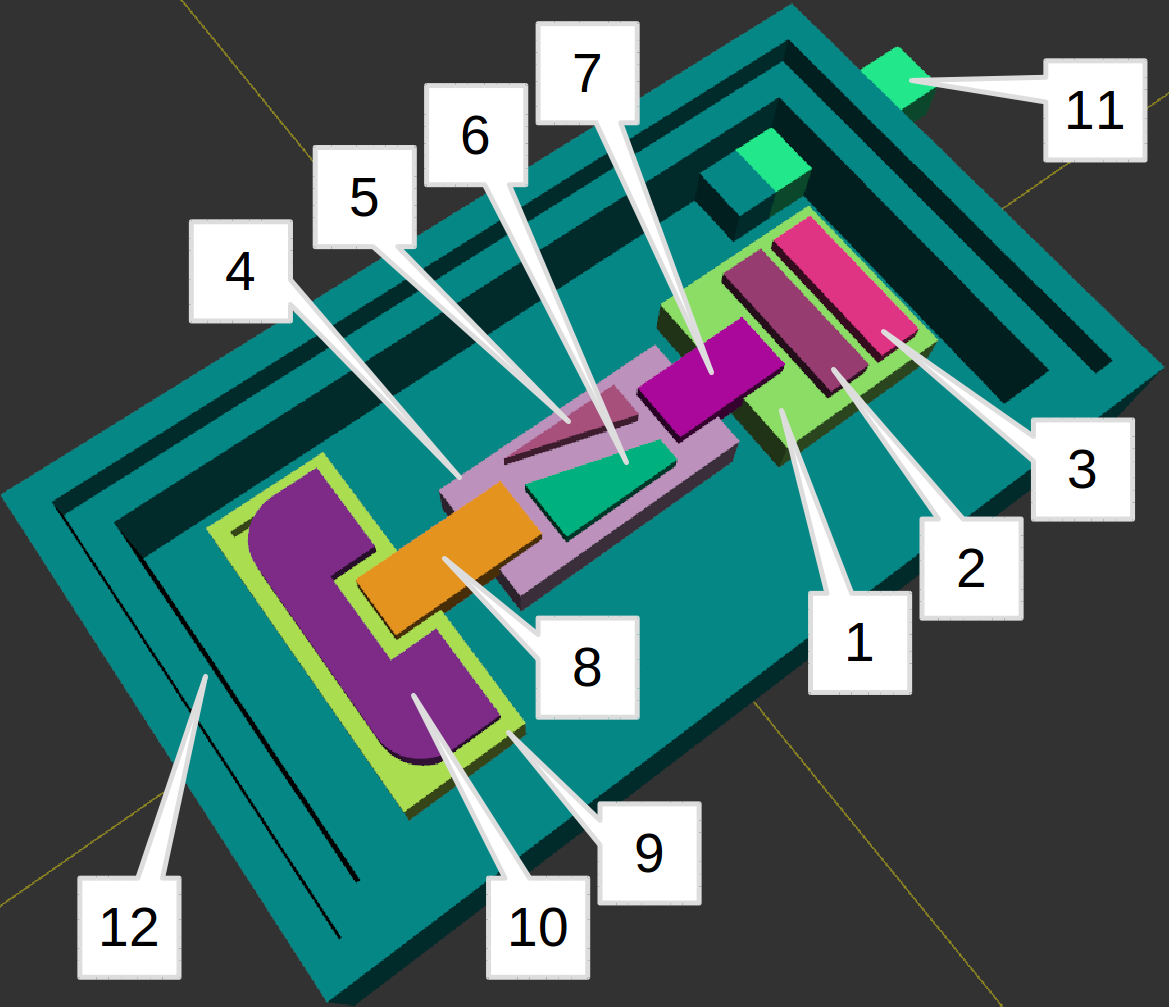}
	\caption{Module Box test case, created for this experiment}
	\label{moduleBoxTopDown}
\end{figure}
\begin{figure}[t]
	\centering
	\includegraphics[scale=\caseScale]{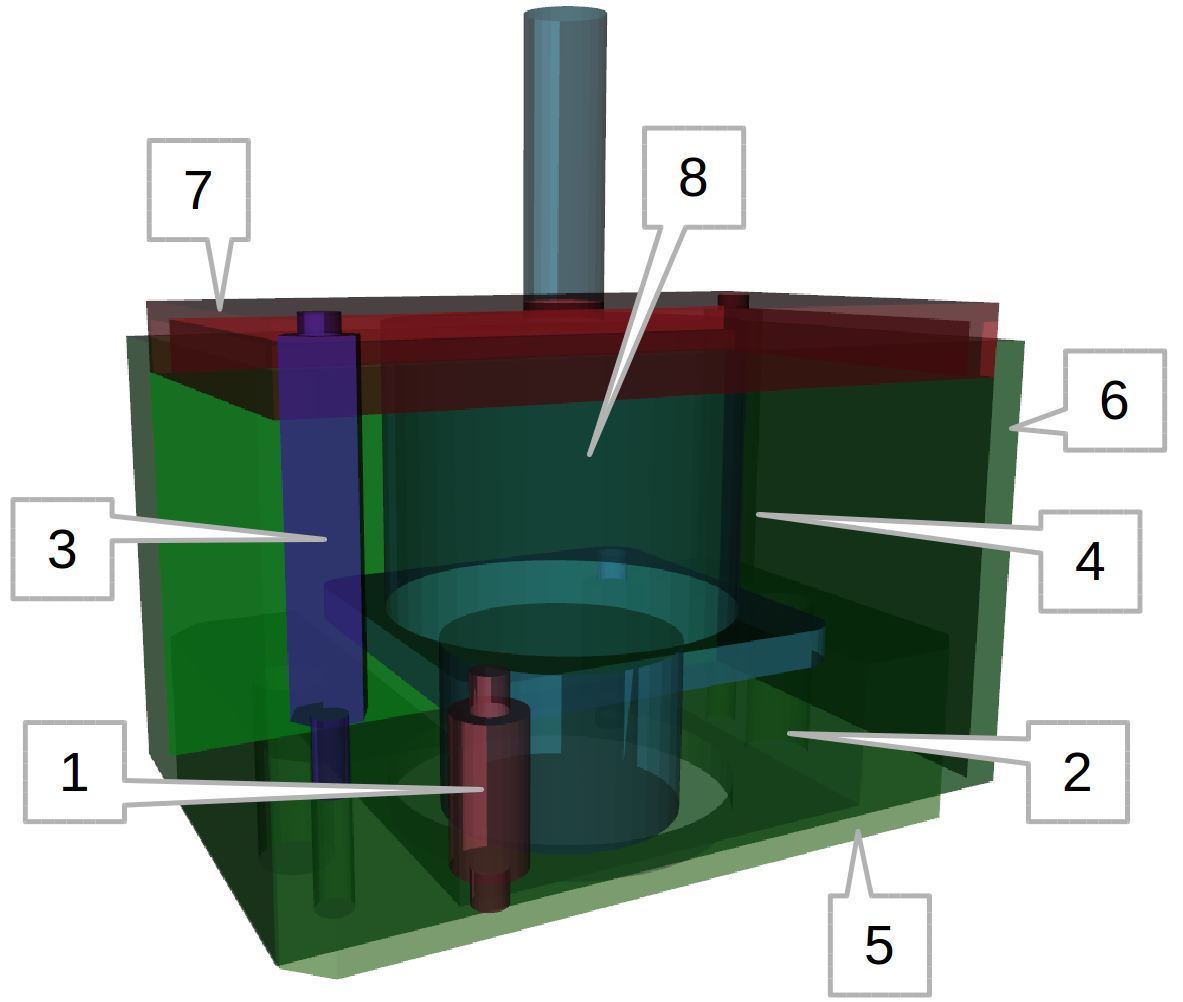}
	\caption{Motor Driver test case, from Wang, Liu, and Zhong \cite{antColony_DRIVEMOTOR}, shown here semi-transparently to reveal the interior parts. A peg and a post are in front of the motor, there are one each of the peg and post $\pi$ radians from those in the foreground.}
	\label{motorTransBig}
	\vspace{-3.5mm}
\end{figure}

\subsection{Path Planning for Action Validation} \label{validate}

Action validation proceeds in three stages.  In the first stage, the local freedom of all parts in the sub-assembly is checked.  If the freedom allowance of the assembly is empty (locked), then the action validation fails.  This happens during subassembly identification, as detailed above.  The second stage is a search for a free translation path through task space in order to determine whether a certain action is feasible, considering collision with other parts and the table on which the assembly rests.

\subsection{Assumptions and Constraints} \label{SECT_assume}

\noindent We make the following assumptions:

\begin{enumerate}
	\item All parts are represented by closed, rigid polyhedral meshes with nominal dimensions. Tolerances are not modeled.
	
	\item Sequential, one-handed operations are assumed. \label{asmp2}
	
	\item All assembly operations are monotonic.  Once joined, two parts remain rigidly joined.
	
	\item All assembly operations are reversible, and a valid removal operation implies that the reverse assembly operation is valid.
	
	\item Adjacent parts in the design are assumed to be connected and can support any other adjacent part. 
	
	\item  Contact dynamics are not modeled.  
\end{enumerate}

We believe these assumptions are reasonable because most manufactured products contain a subset of rigid parts that are immobile relative to one another.

\section{Experiments} \label{SECT_experiment}
In order to show the efficacy of our method, we generate disassembly plans using all three methods described above for two assembly problems.  (Figure \ref{moduleBoxTopDown} and \ref{motorTransBig}) We compare the disassembly plans generated by our method to the plans generated by the methods presented in \cite{kmeansRRT45} and \cite{volumeAreaFitness2} in terms of makespan and speedup.  The products tested are listed in Table \ref{testedAsm}.
\begin{figure}[h!]
	\centering
	\captionof{table}{Tested Products} 	\label{testedAsm}
	\begin{tabular}{|l|r|l|}
		\hline
		\textbf{Design}     & \textbf{Parts}        & \textbf{Source} \\ \hline
		Module Box          & 12                    & N/A         \\ \hline
		Drive Motor         &  8                    & Wang, Liu, \& Zhong \cite{antColony_DRIVEMOTOR}                 \\ \hline
	\end{tabular} 
	\vspace{-4.5mm}
\end{figure}

\subsection{Tested Assemblies}
We used two test cases to compare the methods.  The first is a ``Module Box'', designed for this work.  (Figure \ref{moduleBoxTopDown})   It consists of a larger outer case with a collection of parts inside and an `antenna' structure that protrudes from the top.  It is an assembly with groupings of parts that can form subassemblies.   Note that these groupings are not defined as part of the assembly definition, these must be identified by the methods.  There are also `bridge' structures that span the groupings that require precedence to be considered in order to execute the assembly correctly.  The second is the ``Motor Driver'', presented by Wang, Liu, and Zhong \cite{antColony_DRIVEMOTOR}, that represents an electric motor in an enclosure.  This assembly has parts covering the top and bottom that obstruct the removal of interior parts.

We express the plans produced by each method as tree structures, with the root as the completed assembly, branching into nodes representing subassemblies, and terminating in leaves representing individual parts. 

\subsection{Part Interaction Clusters, Morato 2013} \label{moratoMethod}
Morato et al. \cite{kmeansRRT45} make the assumption that parts that are clustered in physical space are related, and thus candidates for inclusion in a subassembly.  This method was chosen because it is a fairly recent method that employs a simple clustering algorithm to produce candidate subassemblies.  This is an appropriate baseline as the most straightforward means of producing feasible subassemblies.

During the implementation of this method, clusters that were not fully connected were noted.  These obviously cannot be removed as a unit.  The remedy for this was to automatically break up disjoint clusters into connected subassemblies and proceed as though they had been created during the clustering step.  The original work did not address this scenario.

\subsection{Subassembly Generation from CAD, Belhadj 2017} \label{belhadjMethod} 
Belhadj et al. \cite{volumeAreaFitness} also rely on the identification of base parts as the first step of subassembly identification.  Subassemblies accrue neighbors by evaluating their connecting surfaces.  They continue this work with \cite{volumeAreaFitness2} with a different collection of assembly problems.  This method was chosen because it is representative of state of the art methods that use the connectivity and geometric relationships between parts to identify subassemblies.

\subsection{Validation Process} \label{process} 
For the tested assembly problem, the product begins in the fully assembled state.  The environment for the simulation is a flat surface on which the fully assembled product rests.   

For all methods, we enforce constraints on disassembly validation in order to simulate the aspects of a removal operation that are, in general, relevant to the execution of the operation by a robot, but not particular to any model of robot.  The system enforces that the assembly be stably supported under gravity by the floor before and after the removal of the subassembly or parts.  A removal operation is successful if a straight-line removal path can be found that collides neither with the floor nor the reference parts.  We believe this approximation is reasonable because it ensures the current moving subassembly is accessible at its disassembly step, and that the plan produced can be achieved by simple, straight-line movements that are easily executed by widely-available dexterous arms.

Each plan generated by each method is validated as described above.  If a removal action is blocked by the resting surface, then a reorientation action is taken that brings the removal direction closest to vertical.  If validation fails, then the method has failed to produce a correct plan for the assembly.  The consideration of orientation is necessary as part of the validation stage, as we require a stably-supported pose to exist for each step of a plan in order classify it as feasible.
\begin{figure}[b]
	\centering
	\captionof{table}{Performance for Module Box Problem}	\label{makespanModule}
	\begin{tabular}{|l|c|r|r|} \hline
		
		\textbf{Method} & \textbf{Robots} & \textbf{Makespan} & \textbf{Speedup} \\ \hhline{|=|=|=|=|}
		
		\multirow{3}{*}{Block. Reduc.} & 1 & 11 & N/A  \\ \cline{2-4}
									   & 2 &  7 & 1.57 \\ \cline{2-4}
									   & 3 &  6 & 1.83 \\ \hhline{|=|=|=|=|}
		
		\multirow{3}{*}{Morato} & 1 & 11 & N/A  \\ \cline{2-4}
								& 2 &  7 & 1.57 \\ \cline{2-4}
								& 3 &  6 & 1.83 \\ \hhline{|=|=|=|=|}
		
		\multirow{3}{*}{Belhadj} & 1 & 11  & N/A  \\ \cline{2-4}
								 & 2 &  7  & 1.57 \\ \cline{2-4}
							     & 3 &  5  & 2.20 \\ \hline
	\end{tabular} 
	\vspace{-1.5mm}
\end{figure}
\subsection{Assembly Line Simulation} \label{SECT_makespan}
In order to study how plan complexity impacts manufacturing time, we simulate a simple robotic assembly line environment.  The line consists of 1 to 3 worker agents (robots).  Assembly proceeds in discrete time, with one assembly operation (reverse of disassembly) taking one time unit to complete.  Only one subassembly can be assigned to any worker at any time in this scheme.  That is, no two workers ever perform simultaneous actions in which the moved subassemblies have the same reference subassembly.  Each timestep is composed of two phases; assignment and execution.  During assignment, each idle robot is given a job (subassembly) for which the constituents are available: either single free parts or already-completed subassemblies.  (Subassemblies with incomplete constituents are not available to be assigned, thus enforcing the precedence encoded by the tree.)  In the execution phase, each robot marks one action of its job as complete.  If a robot completes its job as a result of the execution step, then that robot becomes idle and the subassembly becomes complete at the start of the next step.  Workpiece rotations and subassembly transfers between workers are not modeled in this scenario for the sake of simplicity.  In this scheme, available subassemblies are assigned in order of decreasing depth in the disassembly graph. 
\begin{figure}[b]
	\centering
	\captionof{table}{Performance for Motor Driver Problem}	\label{makespanMotor}
	\begin{tabular}{|l|c|r|r|} \hline
		
		\textbf{Method} & \textbf{Robots} & \textbf{Makespan} & \textbf{Speedup} \\ \hhline{|=|=|=|=|}
		
		\multirow{3}{*}{Block. Reduc.} & 1 &  7  & N/A \\ \cline{2-4}
									   & 2 &  5  & 1.40 \\ \cline{2-4}
									   & 3 &  5  & 1.40 \\ \hhline{|=|=|=|=|}
		
		\multirow{3}{*}{Morato} & 1 &  7  & N/A \\ \cline{2-4}
								& 2 &  6  & 1.17 \\ \cline{2-4}
								& 3 &  6  & 1.17 \\ \hhline{|=|=|=|=|}
		
		\multirow{3}{*}{Belhadj} & 1 &  7  & N/A \\ \cline{2-4}
								 & 2 &  7  & 1.00 \\ \cline{2-4}
								 & 3 &  7  & 1.00 \\ \hline
	\end{tabular} 
	\vspace{-3.5mm}
\end{figure}
\subsection{Measures for Evaluation} \label{SECT_measureBox}
We use the following measures of disassembly plan efficiency in order to compare the results of each method.  For each measure, the calculation method and the motivating rationale are given. 

The first measure is \textit{makespan}, which is a common in measuring scheduling efficiency. \cite{makeSpan_speedUp,modernMetrics2018}  It is the amount of time between the beginning and the completion of a plan.

The second measure is \textit{speedup}, which is the unit-less ratio of makespan of a plan with no parallelization (serial execution) to the makespan the same plan with parallel execution.  Its aim is to express the multiplication of throughput as a result of adding more workers.  \cite{makeSpan_speedUp}  

\section{Results}
\subsection{Subassembly Decomposition Results} 
Complete disassembly plans were generated for the assembly test cases described in Section \ref{SECT_experiment}.  

\begin{figure}[h!]
	\centering
	\includegraphics[scale=\slideScale]{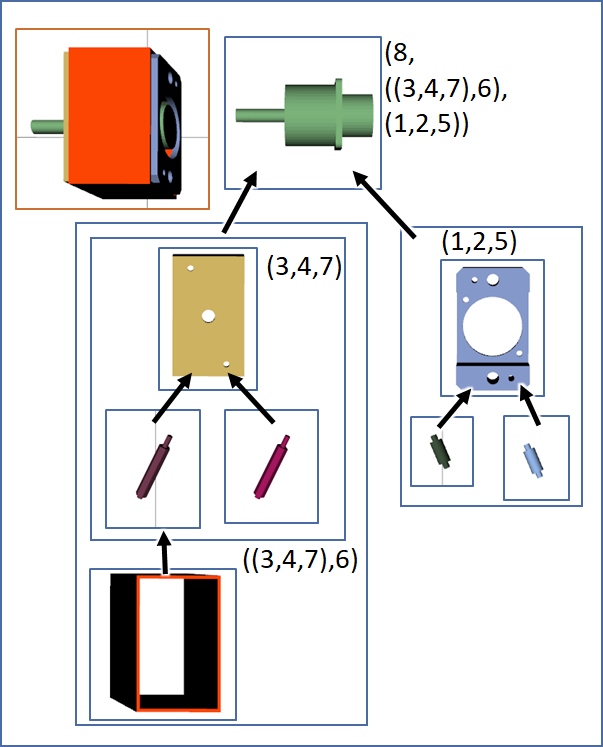}
	\caption{Blocking Reduction assembly plan for the Motor Driver}
	\label{Result_Motor_BR}
	\vspace{-5.5mm}
\end{figure}

\begin{figure}[h!]
	\centering
	\includegraphics[scale=\slideScale]{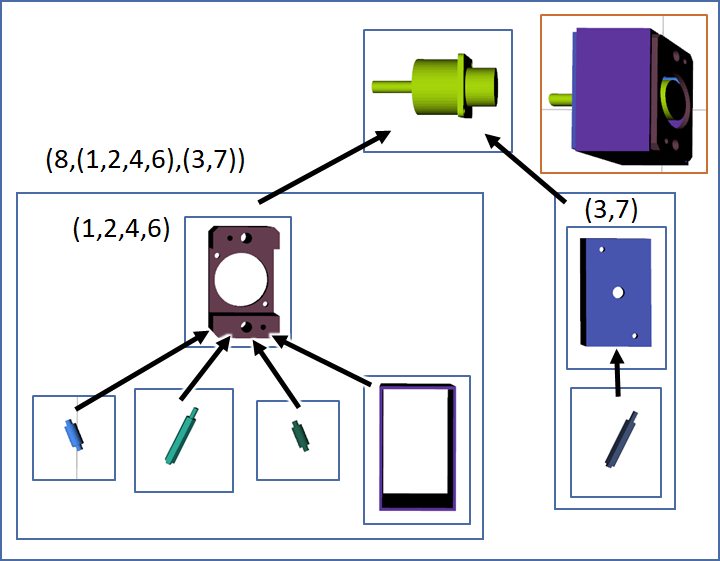}
	\caption{Morato assembly plan for the Motor Driver}
	\label{Result_Motor_Morato}
	\vspace{-6.0mm}
\end{figure}

\begin{figure}[h!]
	\centering
	\includegraphics[scale=\slideScale]{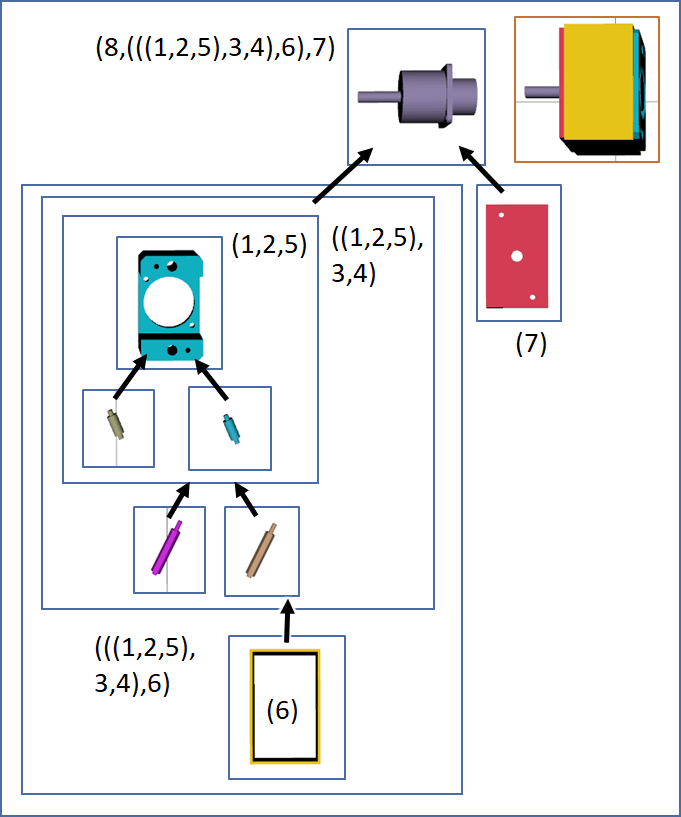}
	\caption{Belhadj assembly plan for the Motor Driver}
	\label{Result_Motor_Belhadj}
	\vspace{-5.5mm}
\end{figure}

The resulting plans are visualized in Figures \ref{Result_Module_BR} through \ref{Result_Motor_Belhadj}.  In each of the plan visualizations, subassemblies are enclosed by a blue box.  Each arrow represents an assembly action with the moving part at the tail and the reference (stationary base) part at the head.  Each stage of disassembly is labeled with the numbers of the constituent parts and subassemblies.  Parentheses indicate subassembly groupings, with a primary subassembly in a single set of parentheses. Each planner was run 10 times on each test case on a computer with a i3-6100 CPU.  For the Module Box, the Morato, Belhadj, and DIG planners ran for an average of 1.65, 1.39, and 10.05 seconds, respectively.  For the Motor Driver, running times were 14.94, 15.52, and 47.15 for the same planners.
 
\subsection{Assembly Line Results} \label{SECT_robotResult}
For each of the plans generated by the tested methods, we ran a simulated assembly procedure as described in Section \ref{SECT_makespan}.  The results of sequence execution for the Module Box and Motor Driver can be seen in Tables \ref{makespanModule} and \ref{makespanMotor}, respectively.  Performance of the sequence execution is expressed in the metrics defined in Section \ref{SECT_measureBox}.
\begin{figure}[t]
	\centering
	\includegraphics[scale=\slideScale]{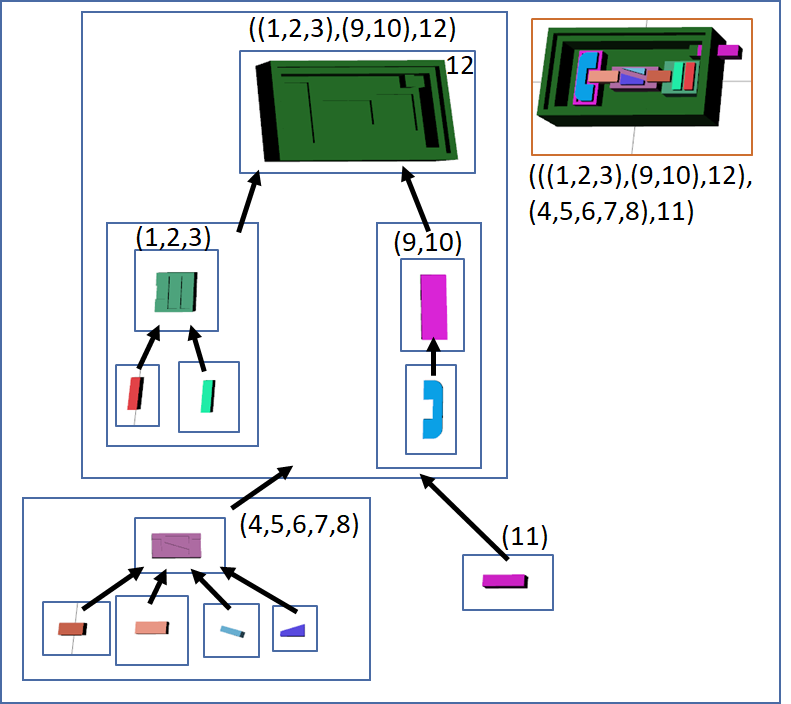}
	\caption{Blocking Reduction assembly plan for the Module Box.}
	\label{Result_Module_BR}
\end{figure}
\begin{figure}[b]
	\centering
	\includegraphics[scale=\slideScale]{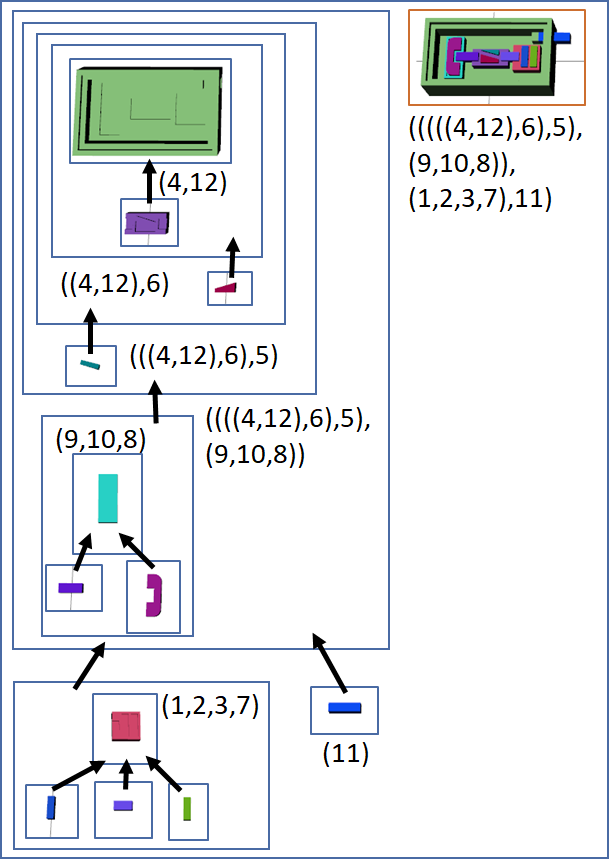}
	\caption{Morato assembly plan for the Module Box.}
	\label{Result_Module_Morato}
	\vspace{-5.5mm}
\end{figure}
Our method and Morato perform identically on the Module Box test case, as seen in Table \ref{makespanModule}.  (Figures \ref{Result_Module_BR} and \ref{Result_Module_Morato})  The plan generated by the Belhadj method has a shorter makespan in the three-robot scenario with a speedup of 2.2 versus the 1.8 speedup for the other two methods.  Belhadj performs identically to the other two methods in and 1 and 2 robot scenarios.

\begin{figure}[h!]
	\centering
	\includegraphics[scale=\slideScale]{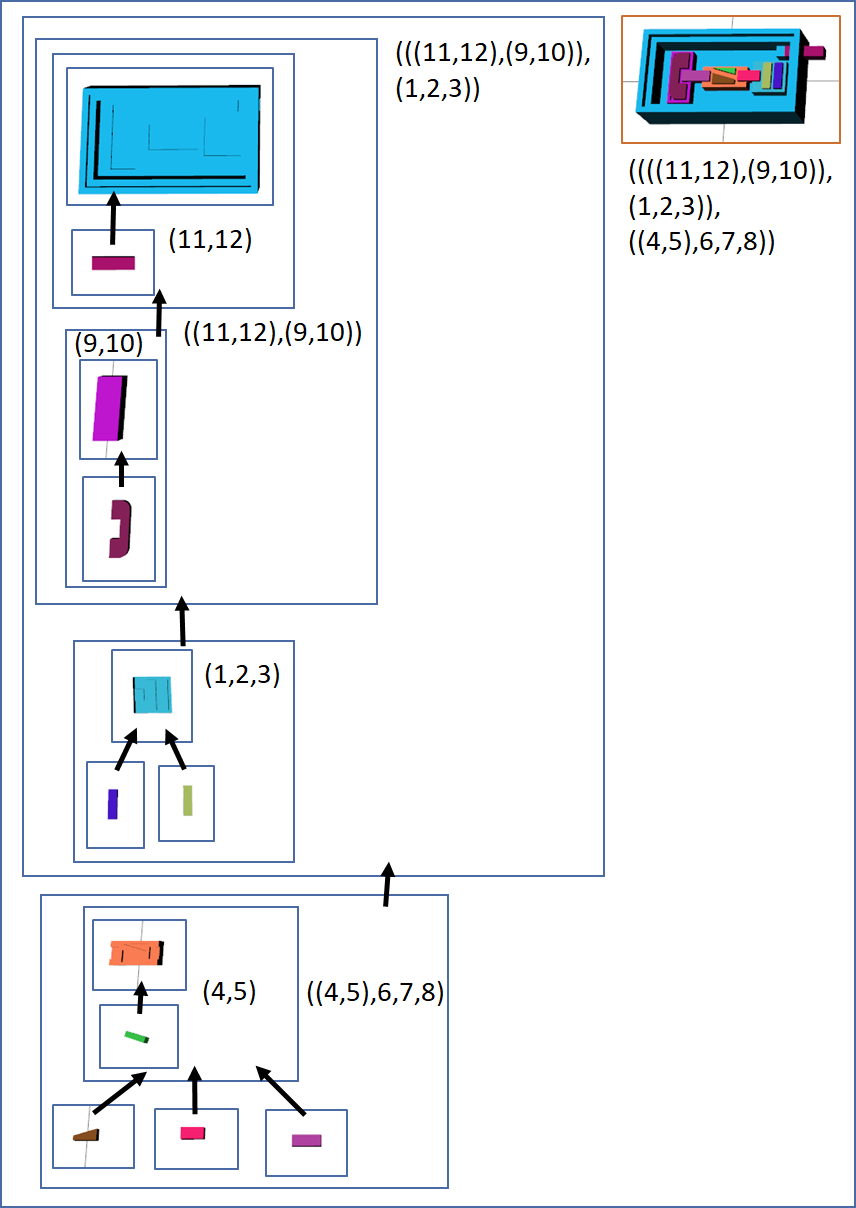}
	\caption{Belhadj assembly plan for the Module Box.}
	\label{Result_Module_Belhadj}
	\vspace{-2.5mm}
\end{figure}

The Belhadj plan for the Motor Driver test case (Table \ref{makespanMotor}) cannot be parallelized at all because it has only a single chain of subassemblies between it and the root, with no parallel branches.  (See Figure \ref{Result_Module_Belhadj}.)  Morato fares slightly better, offering a 1 time-step reduction in makespan because its disassembly tree has 2 branches.  Our method has a speedup of 1.40 versus 1.17 for Morato.  None of the tested methods produce a plan suitable for execution on 3 robots because there are not enough parallel branches to distribute between workers.

The workcell simulation shows that in order for a plan to make the greatest use of parallelization (speedup), it should be composed of the greatest number of branches of equal length as possible.

\section{Conclusion} \label{SECT_conclusion}
We present a graph structure representing the total part-vs.-part blocking state within a total assembly: the disassembly interference graph.  In this structure, removal obstruction is expressed as a scalar heuristic.  We compute this heuristic without undertaking an exhaustive search of removal paths.  We validate these novel contributions by developing a disassembly planner that yields intuitive, tightly-grouped subassemblies.  The performance of our method is comparable to or better than two recent methods in terms of makespan and speedup through parallelization.  The results show that a numeric measure of the obstruction state of parts and subassemblies can be used to generate viable assembly plans, and that consideration of part access must be considered at every stage of planning, not just as a validation step.

\bibliographystyle{ieeetr}
\bibliography{blockReduc.bib} 

\begin{thebibliography}{10}

\bibitem{simplifiedGenAllSeq_GEARTRAIN}
T.~De~Fazio and D.~Whitney, ``Simplified generation of all mechanical assembly
  sequences,'' {\em IEEE Journal on Robotics and Automation}, vol.~3, no.~6,
  pp.~640--658, 1987.

\bibitem{PostProcessingRafiCampbell}
N.~Rafibakhsh and M.~I. Campbell, ``Beyond optimal sequencing: Defining part
  orientation and worker allocation in assembly,'' in {\em ASME 2015
  International Design Engineering Technical Conferences and Computers and
  Information in Engineering Conference}, pp.~V004T05A009--V004T05A009,
  American Society of Mechanical Engineers, 2015.

\bibitem{plansForRobotExec}
W.~Wan, F.~Lu, Z.~Wu, and K.~Harada, ``Teaching robots to do object assembly
  using multi-modal 3d vision,'' {\em Neurocomputing}, vol.~259, pp.~85--93,
  2017.

\bibitem{IKEA}
R.~A. Knepper, T.~Layton, J.~Romanishin, and D.~Rus, ``Ikeabot: An autonomous
  multi-robot coordinated furniture assembly system,'' {\em International
  Conference on Robotics and Automation (ICRA)}, 2016.

\bibitem{kmeansRRT45}
C.~Morato, K.~N. Kaipa, and S.~K. Gupta, ``Improving assembly precedence
  constraint generation by utilizing motion planning and part interaction
  clusters,'' {\em Computer-Aided Design}, vol.~45, no.~11, pp.~1349--1364,
  2013.

\bibitem{volumeAreaFitness2}
M.~Trigui, I.~Belhadj, and A.~Benamara, ``Disassembly plan approach based on
  subassembly concept,'' {\em The International Journal of Advanced
  Manufacturing Technology}, vol.~90, no.~1-4, pp.~219--231, 2017.

\bibitem{liaisonModern}
Y.~Laili, F.~Tao, D.~T. Pham, Y.~Wang, and L.~Zhang, ``Robotic disassembly
  re-planning using a two-pointer detection strategy and a super-fast bees
  algorithm,'' {\em Robotics and Computer-Integrated Manufacturing}, vol.~59,
  pp.~130--142, 2019.

\bibitem{validationBlock_kheder2017}
M.~Kheder, M.~Trigui, and N.~Aifaoui, ``Optimization of disassembly sequence
  planning for preventive maintenance,'' {\em The International Journal of
  Advanced Manufacturing Technology}, vol.~90, no.~5-8, pp.~1337--1349, 2017.

\bibitem{clumpNDBG}
N.~Ong and Y.~Wong, ``Automatic subassembly detection from a product model for
  disassembly sequence generation,'' {\em The international journal of advanced
  manufacturing technology}, vol.~15, no.~6, pp.~425--431, 1999.

\bibitem{volumeAreaFitness}
I.~Belhadj, M.~Trigui, and A.~Benamara, ``Subassembly generation algorithm from
  a cad model,'' {\em The International Journal of Advanced Manufacturing
  Technology}, vol.~87, no.~9-12, pp.~2829--2840, 2016.

\bibitem{AsmSizeComplexity}
R.~M. Marian, {\em Optimisation of assembly sequences using genetic
  algorithms}.
\newblock PhD thesis, University of South Australia, 2003.

\bibitem{subComplexity}
L.~E. Kavraki and M.~N. Kolountzakis, ``Partitioning a planar assembly into two
  connected parts is np-complete,'' {\em Inf. Process. Lett.}, vol.~55,
  pp.~159--165, Aug. 1995.

\bibitem{forceFlowDisasm}
S.~Lee and H.~Moradi, ``Disassembly sequencing and assembly sequence
  verification using force flow networks,'' in {\em Robotics and Automation,
  1999. Proceedings. 1999 IEEE International Conference on}, vol.~4,
  pp.~2762--2767, IEEE, 1999.

\bibitem{maxFlowMinCut}
G.~B. Dantzig and D.~Fulkerson, ``On the min cut max flow theorem of
  networks,'' {\em Annals of Mathematical Study}, vol.~38, pp.~215--222, 1956.

\bibitem{CollsnCheckSubsOnly}
R.~Vigan{\`o} and G.~O. G{\'o}mez, ``Automatic assembly sequence exploration
  without precedence definition,'' {\em International Journal on Interactive
  Design and Manufacturing (IJIDeM)}, vol.~7, no.~2, pp.~79--89, 2013.

\bibitem{SingleArmAsmRegraspWanHarada}
W.~Wan and K.~Harada, ``Integrated single-arm assembly and manipulation
  planning using dynamic regrasp graphs,'' in {\em 2016 IEEE International
  Conference on Real-time Computing and Robotics (RCAR)}, pp.~174--179, June
  2016.

\bibitem{SandiaModularFixture}
R.~G. Brown and R.~C. Brost, ``A 3-d modular gripper design tool,'' {\em IEEE
  Transactions on Robotics and Automation}, vol.~15, no.~1, pp.~174--186, 1999.

\bibitem{FxtrSocialAutoBody}
Y.~Xing and Y.~Wang, ``Assembly operation optimization based on social
  radiation algorithm for autobody,'' {\em Advances in Mechanical Engineering},
  vol.~6, p.~854637, 2014.

\bibitem{SupportOnPinsWanHarada}
C.~Cao, W.~Wan, J.~Pan, and K.~Harada, ``Analyzing the utility of a support pin
  in sequential robotic manipulation,'' in {\em Robotics and Automation (ICRA),
  2016 IEEE International Conference on}, pp.~5499--5504, IEEE, 2016.

\bibitem{freedomCone}
B.~Romney, {\em Atlas: An Automatic Assembly Sequencing and Fixturing System},
  pp.~397--415.
\newblock Berlin, Heidelberg: Springer Berlin Heidelberg, 1997.

\bibitem{projectionCone2p5D}
U.~Thomas, M.~Barrenscheen, and F.~M. Wahl, ``Efficient assembly sequence
  planning using stereographical projections of c-space obstacles,'' in {\em
  Assembly and Task Planning, 2003. Proceedings of the IEEE International
  Symposium on}, pp.~96--102, IEEE, 2003.

\bibitem{RAPID}
G.~Zachmann, ``Rapid collision detection by dynamically aligned dop-trees,'' in
  {\em Proceedings. IEEE 1998 Virtual Reality Annual International Symposium
  (Cat. No. 98CB36180)}, pp.~90--97, IEEE, 1998.

\bibitem{antColony_DRIVEMOTOR}
J.~Wang, J.~Liu, and Y.~Zhong, ``A novel ant colony algorithm for assembly
  sequence planning,'' {\em The international journal of advanced manufacturing
  technology}, vol.~25, no.~11, pp.~1137--1143, 2005.

\bibitem{makeSpan_speedUp}
T.~Hagras and J.~Janecek, ``A high performance, low complexity algorithm for
  compile-time task scheduling in heterogeneous systems,'' in {\em Parallel and
  Distributed Processing Symposium, 2004. Proceedings. 18th International},
  p.~107, IEEE, 2004.

\bibitem{modernMetrics2018}
J.~A. Marvel, R.~Bostelman, and J.~Falco, ``Multi-robot assembly strategies and
  metrics,'' {\em ACM Computing Surveys (CSUR)}, vol.~51, no.~1, p.~14, 2018.

\end{thebibliography}

\end{document}